\documentclass{article}
\usepackage{spconf,amsmath,graphicx}
\usepackage{amsfonts}
\usepackage{float}

\def\4DVAR{4DVAR}
\title{Learning 4DVAR inversion directly from observations}

\name{Arthur Filoche${}^1$, Julien Brajard${}^2$, Anastase Charantonis ${}^3$, Dominique Béréziat${}^1$}

\address{Sorbonne Université, CNRS, LIP6, France${}^1$ \\
Sorbonne Université, CNRS, LOCEAN, France${}^2$ 
\\ ENSIIE, CNRS, LAMME, France${}^3$}
  
\begin{document}
%
\maketitle
\begin{abstract}
Variational data assimilation and deep learning share many algorithmic aspects in common. While the former focuses on system state estimation, the latter provides great inductive biases to learn complex relationships. We here design a hybrid architecture learning the assimilation task directly from partial and noisy observations, using the mechanistic constraint of the \4DVAR algorithm. Finally, we show in an experiment that the proposed method was able to learn the desired inversion with interesting regularizing properties and that it also has computational interests.
\end{abstract}
\begin{keywords}
Data Assimilation, Unsupervised Inversion, Differentiable Physics, Learning Simulation
\end{keywords}

\section{Introduction}\label{sec:intro}

Data Assimilation~\cite{carrassi2018data} is a set of statistical methods solving particular inverse problems, involving a dynamical model and imperfect data obtained through an observation process, with the objective to estimate a considered system state. It produces state-of-the-art results in various numerical weather prediction tasks and is mostly used in operational meteorological centers. 

Although they are not initially designed for the same purpose, variational data assimilation~\cite{Tellus} and deep learning share many algorithmic aspects~\cite{abarbanel18}. It has already been argued that both methods can benefit from each other~\cite{nature,geer}. Data assimilation provides a proper Bayesian framework to combine sparse and noisy data with physics-based knowledge while deep learning can leverage a collection of data extracting complex relationships from it. Hybrid methods have already been developed either to correct model error~\cite{correct_model_error,duben2021machine}, to jointly estimate parameters and system state~\cite{Bocquet_2020,nguyen:hal-02436060} or to fasten the assimilation process~\cite{FDA}. Most of these algorithms rely on iterative optimization schemes alternating data assimilation and machine learning steps. 

In this work we design a hybrid architecture bridging a neural network and a mechanistic model to directly learn system state estimation from a collection of partial and noisy observations. We optimize it in only one step still using the variational assimilation loss function. Finally, We show in an experiment using the chaotic Lorenz96 dynamical system, that the proposed method is able to learn the variational data assimilation with desirable regularizing properties, then providing a computationally efficient inversion operator.

\section{Related work}

\subsection{Hybridizing data assimilation with machine learning}\label{sec:r_work}

While deep learning has proven to be extremely useful for a variety of inverse problems where the ground truth is available, unsupervised inversion is still being investigated~\cite{Willet}. For instance, when data are highly-sparse, neural architectures may be hard to train. On the other hand, data assimilation can provide dense data. From this statement, approaches have naturally emerged in the data assimilation community, iterating data assimilation steps and machine learning steps for simultaneous state and parameters estimation~\cite{Bocquet_2020,nguyen:hal-02436060}. But end-to-end learning approaches are also investigated, in~\cite{Fablet_end2end} the architecture is constrained to internally behave like a \4DVAR pushing the hybridization further.

\subsection{Mechanistically constrained neural networks}

Variational data assimilation has a pioneering expertise in PDE-constrained optimization~\cite{Tellus}, making use of automatic differentiation to retro-propagate gradients through the dynamical system. In~\cite{Pajot, mosser2018stochastic} the output of a neural network is used as input in a  dynamical model, and architectures are trained with such gradients, in a supervised and adversarial manner, respectively. Similar methods have been used to learn accurate numerical simulations still using differentiable mechanistic models~\cite{acceleratingEulerian,diffPhysics}. Also, Physically-consistent architectures are developed to enforce the conservation of desired quantity by neural architectures~\cite{Rasp}.

\section{Data assimilation and learning framework}\label{sec:DAL}

\subsection{State-space system}

A system state $\mathbf{X}_{t}$ evolves over time according to a considered perfectly known dynamics $\mathbb{M}_{t}$ and observations $\mathbf{Y}_{t}$ are obtained through an observation operator $\mathbb{H}$ up to an additive noise $\varepsilon_{R_t}$, as described in Eqs.~\ref{eq:M} and~\ref{eq:H},

\begin{align}
&\text{Dynamics:} \quad &\mathbf{X}_{t+1}& = \mathbb{M}_t(\mathbf{X}_t)\label{eq:M}\\ 
&\text{Observation:} \quad &\mathbf{Y}_t& = \mathbb{H}_t(\mathbf{X}_t) + \varepsilon_{R_t} \label{eq:H}
\end{align}

We denote the trajectory $\mathbf{X}=[\mathbf{X}_0,\dots,\mathbf{X}_T]$, a sequence of state vectors over a temporal window, and $\mathbf{Y}$ the associated observations. The objective of data assimilation is to provide an estimation of the posterior probability $p(\mathbf{X}\mid \mathbf{Y})$ leveraging the information about the mechanistic model $\mathbb{M}$. The estimation can later be used to produce a forecast.

\subsection{The initial value inverse problem}
When considering the dynamics perfect, the whole trajectory only depends on the initial state $\mathbf{X}_0$, the assimilation is then said with strong-constraint. The whole process to be inverted is summed up in the simple Eq.~\ref{eq:forward_inverse}, where $\mathcal{F}$ is the forward model, combining $\mathbb{M}_t$ and $\mathbb{H}_t$. More precisely, by denoting multiple model integrations between two times $\mathbb{M}_{t_1 \to t_2}$, we can rewrite the observation equation as in Eq.~\ref{eq:forward_t}.

\begin{equation}
\label{eq:forward_inverse} 
\mathbf{Y} = \mathcal{F}(\mathbf{X}_0) + \varepsilon_{R}
\end{equation}

\begin{equation}
\label{eq:forward_t} 
\mathbf{Y}_t  = \mathbb{H}_t \circ \mathbb{M}_{0 \to t} (\mathbf{X}_0) + \varepsilon_{R_t}
\end{equation}

The desired Bayesian estimation now requires a likelihood model $p(\mathbf{X}\mid\mathbf{Y})$ and a prior model $p(\mathbf{X})=p(\mathbf{X}_0)$. We assume the observation errors uncorrelated in time so that $p(\mathbf{X}\mid\mathbf{Y})=\prod_{t}p(\varepsilon_{R_t})$ and we here make no particular assumption on $\mathbf{X}_0$ corresponding to a uniform prior.

\subsection{Variational assimilation with 4DVAR}
The solve this problem in a variational manner, it is convenient to also assume white and Gaussian observational errors $\varepsilon_{R_t}$, of know covariance matrices $\mathbf{R}_t$, leading to the least-squares formulation given in Eqs.~\ref{eq:4DV_posterior}, where $\|\varepsilon_{R_t}\|_{R_t}^2$ stands for the Mahalanobis distance associated with the matrix $\mathbf{R}_t$. The associated loss function is denoted $\mathcal{J}_{4DV}$ (see Eq.~\ref{eq:j_4dv}) and minimizing it corresponds to a maximum a posteriori estimation, here equivalent to a maximum likelihood estimation.


\begin{equation}
\begin{aligned}
-\log p(\mathbf{X}\mid\mathbf{Y}) & = \frac{1}{2}\sum_{t=0}^{T} \|\varepsilon_{R_t}\|_{\mathbf{R}_t}^2 -\log K\\
& \text{s.t. } \mathbb{M}(\mathbf{X}_t)=\mathbf{X}_{t+1}\\
\end{aligned}
\label{eq:4DV_posterior}
\end{equation}

\begin{equation}
\label{eq:j_4dv} 
\mathcal{J}_{4DV}(\mathbf{X}_0) = \frac{1}{2}\sum_{t=0}^{T} \|\mathbb{H}_t \circ \mathbb{M}_{0 \to t} (\mathbf{X}_0)-\mathbf{Y}_t\|_{\mathbf{R}_t}^2  
\end{equation}

This optimization is an optimal control problem where the initial state $\mathbf{X}_0$ plays the role of control parameters. Using the adjoint state method, we can derive an analytical expression of $\nabla_{\mathbf{X}_0} \mathcal{J}_{4DV}$ as in Eq.~\ref{eq:adjoint}. It is worth noting that the mechanism at stake here is equivalent to the back-propagation algorithm used to train neural networks. The algorithm associated with this optimization is named \4DVAR.

\begin{equation}
\label{eq:adjoint}
    \nabla_{\mathbf{X}_0} \mathcal{J}_{4DV}(\mathbf{X}_0)
     =\sum_{t=0}^{T}\left[
            \frac{\partial(\mathbb{H}_t \circ \mathbb{M}_{0\to t})}{\partial \mathbf{X}_0}
              \right]^\top{\mathbf{R}_t}^{-1} \varepsilon_{R_t}
\end{equation}

\subsection{Learning inversion directly from observations}

We now consider independent and identically distributed trajectories denoted and the dataset of observations $\mathcal{D} = \{\mathbf{Y}^{(i)},\mathbf{R}^{-1(i)}\}_{i=1}^N$. The associated ground truth $\mathcal{T}=\{\mathbf{X}^{(i)}\}_{i=1}^N$ is not available so the supervised setting is not an option. The posteriors for each trajectory are then also independent as developed in Eq.~\ref{eq:posterior_D}.

\begin{equation}
\label{eq:posterior_D}
    \log p(\mathcal{T}\mid\mathcal{D})=\sum_{i=0}^{N}\log p(\mathbf{X}^{(i)}\mid\mathbf{Y}^{(i)})
\end{equation}

Our objective is to learn a parameterized pseudo-inverse $\mathcal{F}^{\star}_{\boldsymbol \theta}$ that should output initial condition from observations and associated errors covariance (see Eq.~\ref{eq:e2e_f}), which is exactly the task solved by \4DVAR. Such modeling choice corresponds to the prior $p(\mathbf{X}_0)=\delta(\mathbf{X}_0-\mathcal{F}^{\star}_{\boldsymbol \theta}(\mathbf{Y},\mathbf{R}^{-1}))$, as we do not use additional regularization, where $\delta$ is the Dirac measure.

\begin{equation}
\label{eq:e2e_f}
\mathcal{F}^{\star}_{\boldsymbol \theta} : (\mathbf{Y},\mathbf{R}^{-1})  \mapsto  \mathbf{X}_0
\end{equation}

\begin{figure}[htbp]
\centering
\includegraphics[width=8cm]{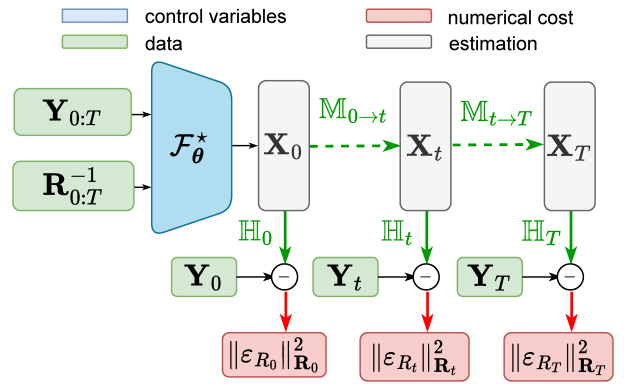}
\caption{Schematic view of the hybrid architecture learning the \4DVAR inversion \label{fig:CompGraph_E2E}}
\end{figure}

To learn the new control parameters $\boldsymbol \theta$, we leverage the knowledge of the dynamical model $\mathbb{M}$ as in \4DVAR. After outputting the initial condition $\mathbf{X}_0$ we forward it with the dynamical model and then calculate the observational loss. A schematic view of the performed integration is drawn in Fig.~\ref{fig:CompGraph_E2E}.

Then the cost function associated with the MAP estimation can be developed as in Eq.~\ref{eq:coste2e}. A simple way of thinking it is to run multiple \4DVAR in parallel to optimize a common set of control parameters $\boldsymbol \theta$.

\begin{equation}
\label{eq:coste2e}
\begin{aligned}
\mathcal{J}({\boldsymbol \theta}) =\sum_{\mathcal{D}} \mathcal{J}_{4DV}(\mathbf{X}^{(i)}_0)\\
\text{s.t. } \mathcal{F}_{\boldsymbol \theta} (\mathbf{Y}^{(i)},\mathbf{R}^{-1(i)})=\mathbf{X}^{(i)}_0\\
\end{aligned}
\end{equation}

To calculate $\nabla_{\boldsymbol \theta}\mathcal{J}$ we simply use the linearity of the gradient (Eq.~\ref{eq:e2egrad2}), then the chain rule (Eq.~\ref{eq:e2egrad2}) and finally we can re-use $\nabla_{\mathbf{X}_0}\mathcal{J}_{4DV}$ calculated before (Eq.~\ref{eq:adjoint}). Gradients are back-propagated through the dynamical model first and then through the parameterized pseudo-inverse. Calculating the gradient on the whole dataset at each iteration may be computationally too expensive so one could instead use mini-batch gradient descent.

\begin{equation}
     \nabla_{\boldsymbol \theta}\mathcal{J}
     =\sum_{\mathcal{D}}\nabla_{\boldsymbol \theta} \mathcal{J}_{4DV}
    \label{eq:e2egrad2}
\end{equation}
\begin{equation}
    \nabla_{\boldsymbol \theta} \mathcal{J}_{4DV}
     =\nabla_{\mathbf{X}_0}\mathcal{J}_{4DV}\nabla_{\boldsymbol \theta}\mathbf{X}_0 = \nabla_{\mathbf{X}_0}\mathcal{J}_{4DV} \nabla_{\boldsymbol \theta}\mathcal{F}^{\star}_{\boldsymbol \theta}
    \label{eq:e2egrad3}
\end{equation}

\section{Experiments and Results}

\subsection{Lorenz96 dynamics and observations}
We use the Lorenz96 dynamics~\cite{L96} as an evolution model Lorenz96 (see Eq.~\ref{eq:Lorezn96}) numerically integrated with a fourth-order Runge Kutta scheme. Here $n$ indexes a one-dimensional space. On the right-hand side, the first term corresponds to an advection, the second term represents damping and $F$ is an external forcing. We use the parameters $dt=0.1$ and $F=8$ corresponding to a chaotic regime \cite{4DLorenz}. Starting from white noise and after integrating during a spin-up period to reach a stationary state, we generate ground truth trajectories.

\begin{equation}
\label{eq:Lorezn96}
     \frac{d\mathbf{X}_{t,n}}{dt}=(\mathbf{X}_{t,n+1}-\mathbf{X}_{t,n-2})\mathbf{X}_{t,n-1}-\mathbf{X}_{t,n}+F
\end{equation} 

\begin{figure}[htbp]
\centering
\includegraphics[width=7cm]{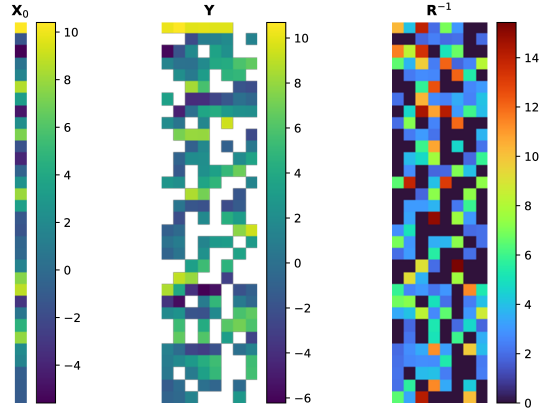}
\caption{Observation generated with the Lorenz96 model, a randomized linear projector as observation operator and a white noise. \label{fig:obs_L96}}
\end{figure}

To create associated observations, we use a randomized linear projector as observation operator, making the observation sparse to finally add a white noise. Noises at each point in time and space can have different variances, $\varepsilon_{R_{n,t}} \sim \mathcal{N}(0,\,\sigma_{n,t})$, and we use the associated diagonal variance matrix defined by $\mathbf{R}_{n,t}^{-1}=\frac{1}{\sigma^{2}_{n,t}}$. Figure~\ref{fig:obs_L96} displays an example of simulated observations. Variances are sampled uniformly such that $\sigma_{n,t} \sim \mathcal{U}(0.25,1)$. When a point in the grid is not observed we fix ``$\mathbf{R}_{n,t}^{-1}=0$'', which corresponds to an infinite variance meaning a lack of information. From a numerical optimization view, no cost means no gradient back-propagated which is the desired behavior.

\subsection{Algorithm benchmarks}

We evaluate our method (NN-\4DVAR-e2e) on the assimilation task which is estimating $\mathbf{X}_0$. We compare it with a \4DVAR, a \4DVAR with additional $\mathcal{L}_2$ regularization (\4DVAR-B), a neural network trained on the output of both \4DVAR estimations (NN-\4DVAR-iter and NN-\4DVAR-B-iter), and a neural network trained with the ground truth (NN-perfect). The latter should represent the best-case scenario for the chosen architecture while NN-\4DVAR-iter plays the role of the iterative method. The same neural architecture is used for all the methods involving learning. Its design is fairly simple, being composed of 5 convolutional layers using $3\times3$ kernels, ReLu activation, no down-scaling, and a last layer flattening the two-dimensional maps into the shape of $\mathbf{X}_0$. We use $250$, $50$, and $250$ samples for training, validation, and testing, respectively. When learning is involved, the Adam optimizer is used while \4DVAR is optimized with the L-BFGS solver. We notice here that once learned, both NN-\4DVAR-iter and NN-\4DVAR-e2e provide a computationally cheap inversion operator. For their learning, the computationally intensive step was the forward integration of the dynamical model. Denoting $n\_iter$ the number of iterations done in \4DVAR and $n\_epoch$ the number of epochs in our learning process, NN-\4DVAR-iter and \4DVAR-e2e cost $N \times n\_iter$ and $N \times n\_epoch$ dynamics integration, respectively. Depending on these parameters, one approach or the other will be less computationally intensive. In our case, we used $n\_iter<150$ and $n\_epoch=50$.

\subsection{Results}

\begin{figure}[htbp]

\begin{minipage}[b]{1.0\linewidth}
  \centering
  \centerline{\includegraphics[width=7.5cm]{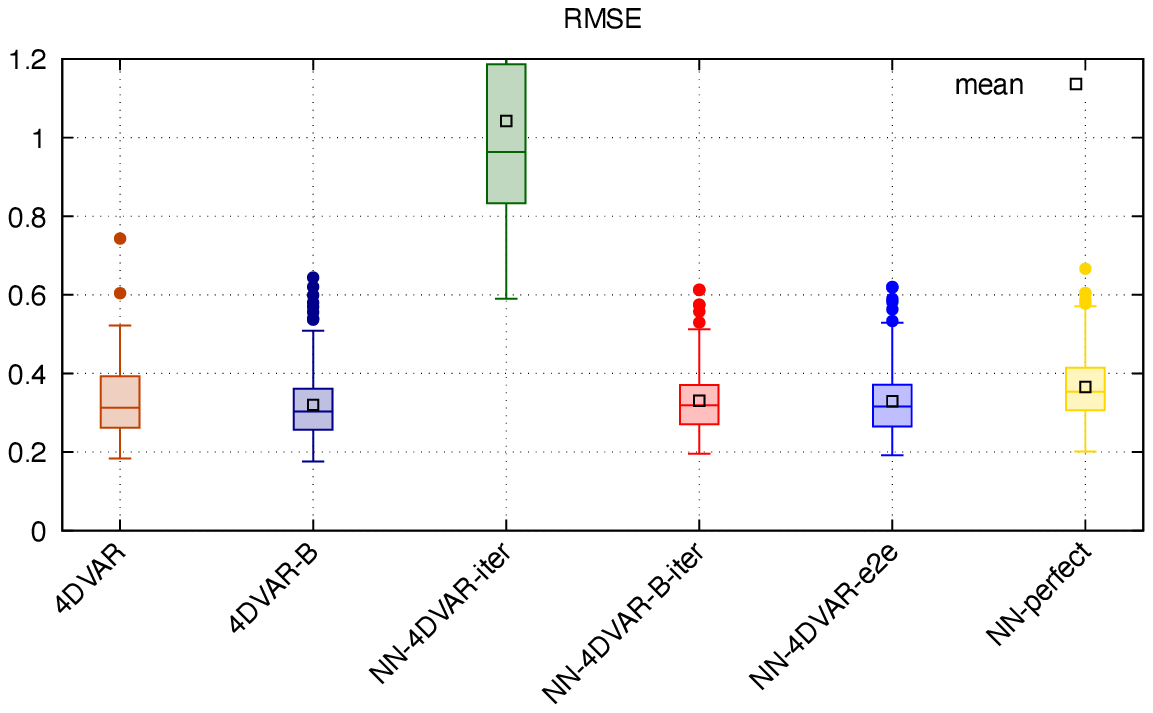}}
\end{minipage}
\begin{minipage}[b]{1.0\linewidth}
  \centering
  \centerline{\includegraphics[width=7.5cm]{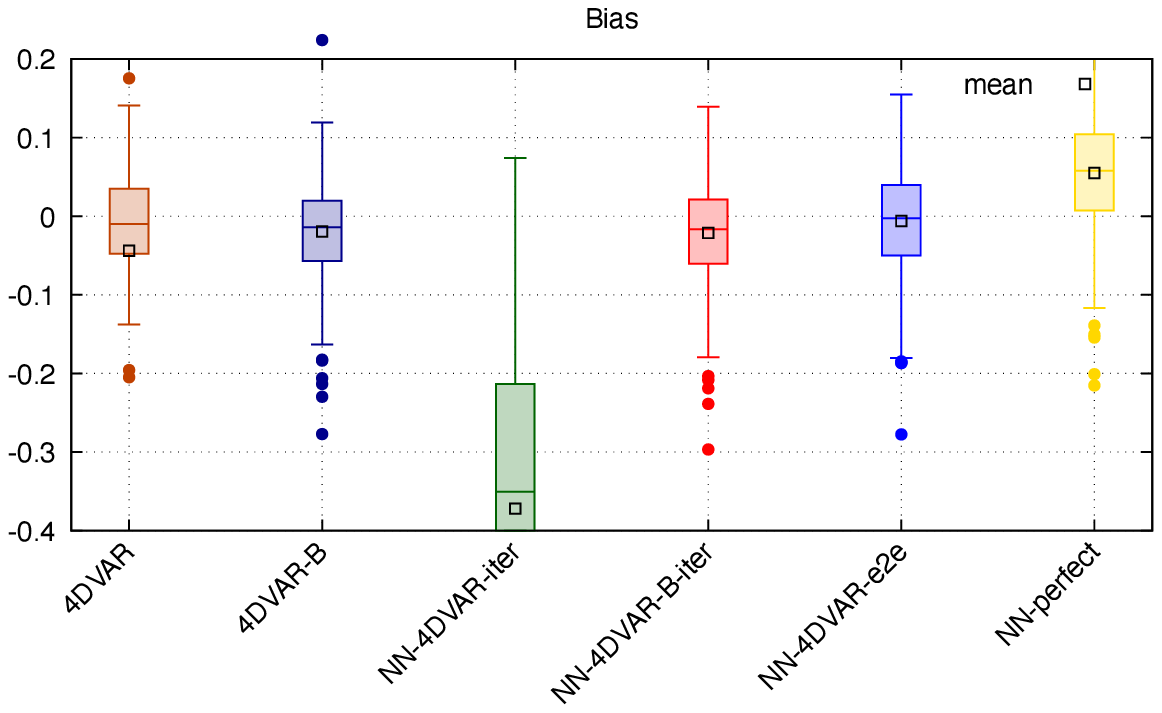}}
\end{minipage}
\caption{Boxplot of assimilation accuracy of each algorithms, RMSE and Bias scores, on the , 250 samples test set}
\label{fig:rmse_bias}
\end{figure}

\begin{figure}[htbp]
\begin{minipage}[b]{1.0\linewidth}
  \centering
  \centerline{\includegraphics[width=6cm]{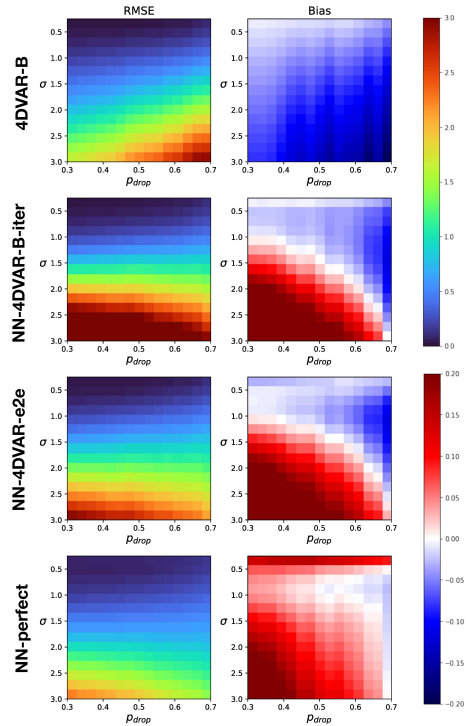}}
\end{minipage}
\caption{Sensitivity of the assimilation regarding noise and sparsity levels ($\sigma$, $p_{drop}$), at each pixel levels are constant and scores are averaged on 25 samples. $\sigma > 1$ not seen in training.}
\label{fig:sensitivity}

\end{figure}

The accuracy of the $\mathbf{X}_0$ estimation 
on the test set is quantified using the RMSE and the average bias (see Figs~\ref{fig:rmse_bias}). We notice first that when \4DVAR is not regularized, some samples induce bad estimations which disturb \4DVAR-NN-iter learning over them. The others methods involving produce RMSE scores on par with \4DVAR-B, the best estimator. However, our \4DVAR-NN-e2e is the less biased algorithm. It is to be noted that \4DVAR-NN-e2e has no additional regularization and still stays robust regarding difficult samples, highlighting desirable properties from the neural architecture. In Fig.~\ref{fig:sensitivity}, we performed an accuracy sensitivity experiment regarding noise and sparsity levels. Particularly, we tested noise levels out of the dataset distribution. We see that learning-based approaches are more sensitive to noise increases while \4DVAR is more concerned by sparsity. Also, we notice that our NN-\4DVAR-e2e methods generalize better than NN-\4DVAR-B-iter to unseen levels of noise.

\section{Conclusion}
\subsection{Overview}
We proposed a hybrid architecture inspired by the \4DVAR algorithm allowing to use of the data assimilation Bayesian framework while leveraging a dataset to learn an inversion operator. We showed in an assimilation experiment that the algorithm was able to desired function while having a stable behavior.

\subsection{Critical discussion}
The designed algorithm fixes the maximum temporal size of the assimilation window. For smaller windows, it can still be used filling the masking variance with zeros accordingly but for larger ones, the only possibility is to use sliding windows, then raising to question of the coherence in time. Typically, the method in that form can not fit quasi-static strategies~\cite{quasi} employed in variational assimilation. Also, We made the convenient hypothesis that observational errors are uncorrelated in space, so that $\mathbf{R}^{-1}$ can be reshaped in the observation format, which may not be the case depending on the sensors. However, the method has a computational interest. Once the parameterized inversion operator learned, the inversion task becomes computationally cheap. But this also stands for the iterative approaches. As discussed before, learning the inversion directly with our method may be less computationally costly, in terms of dynamics integration, depending on the number of epochs when learning our architecture, the number of samples in the dataset, and the number of iterations used in \4DVAR.

\subsection{Perspective}
One of the motivations for the designed architecture was to circumvent algorithms iterating data assimilation and machine learning steps, because of their difficulty of implementation but also their potential bias as exhibited in the experiment. However, we made the debatable, simplifying, perfect model hypothesis. Usually, the forward operator is only partially known and we ambition to develop the proposed framework further to relax such a hypothesis.

\vfill
\pagebreak


\nocite{*}
\bibliographystyle{IEEEbib}
\bibliography{strings,refs}

\end{document}